\documentclass[preprint,review,12pt]{elsarticle}

\usepackage{amssymb}
\usepackage{algorithmicx,algorithm}
\usepackage{hyperref}
\usepackage{multirow}
\usepackage{stfloats}
\usepackage{graphicx}
\usepackage{color}
\usepackage[justification=centering]{caption}
\usepackage{url}
\usepackage{xcolor}
\usepackage{bm}

\journal{Pattern Recognition}

\begin{document}


\begin{frontmatter}

\title{GasHis-Transformer: A Multi-scale Visual Transformer Approach for Gastric Histopathological Image Detection}

\author[1]{Haoyuan Chen}

\author[1]{Chen Li\corref{cor1}}
\cortext[cor1]{Corresponding author:} \ead{lichen201096@hotmail.com}

\author[2]{Ge Wang\corref{cor1}}  
\ead{wangg6@rpi.edu}

\author[3]{Xiaoyan Li}

\author[1]{Md Rahaman}

\author[3]{Hongzan Sun}

\author[1]{Weiming Hu}

\author[1]{Yixin Li}

\author[1]{Wanli Liu}

\author[1,4]{Changhao Sun}

\author[1]{Shiliang Ai}

\author[5]{Marcin Grzegorzek}

\address[1]{Microscopic Image and Medical Image Analysis Group, 
Northeastern University, China}

\address[2]{Department of Biomedical Engineering, Rensselaer Polytechnic Institute, US}

\address[3]{Department of Pathology, China Medical University, China}

\address[4]{Shenyang Institute of Automation, Chinese Academy of Sciences, China}

\address[5]{Institute of Medical Informatics, University of Lübeck, Germany}


\begin{abstract}
In this paper, a multi-scale visual transformer model, referred as GasHis-Transformer, is proposed for \emph{Gastric Histopathological Image Detection} (GHID), which enables the automatic global detection of gastric cancer images. GasHis-Transformer model consists of two key modules designed to extract global and local information using a position-encoded transformer model and a convolutional neural network with local convolution, respectively. A publicly available hematoxylin and eosin (H\&E) stained gastric histopathological image dataset is used in the experiment. Furthermore, a Dropconnect based lightweight network is proposed to reduce the model size and training time of GasHis-Transformer for clinical applications with improved confidence. Moreover, a series of contrast and extended experiments verify the robustness, extensibility and stability of GasHis-Transformer. In conclusion, GasHis-Transformer demonstrates high global detection performance and shows its significant potential in GHID task.
\end{abstract}


\end{frontmatter}

   
\vspace{-0.7cm}
\section{Introduction}
\label{intro}

\vspace{-0.3cm}
Cancer is a malignant tumor that originates from epithelial tissue and is one of the deadliest diseases, which caused approximately 9.6 million deaths in 2018---the highest number since records began in the 1970s. Among all the cancer categories, gastric cancer has the second-highest rate globally in terms of morbidity and mortality. Gastric cancer is a collection of abnormal cells that form tumors in the stomach. In histopathology, the most common type of gastric cancer is adenocarcinoma, which starts in mucous-producing cells in the stomach's inner layer that invade the stomach wall, infiltrating the muscular mucosa and then invade the outer layer. According to World Health Organization statistics, about 800,000 people die due to cancer every year~\cite{hegde2020top}. Therefore, medical staff needs to diagnose gastric cancer accurately and efficiently.

The diagnosis of gastric cancer is performed by carefully examining Hematoxylin and Eosin (H\&E) sections by pathologists under a microscope. This conventional process is time-consuming and subjective. Because of these shortcomings, pathologists face difficulties with accurate screening and diagnosis of gastric cancer. Thus, computer-aided diagnosis (CAD) that began in the 1980s can overcome these shortcomings by making diagnostic decisions with improving efficiency. CAD aims to improve medical doctors' examination quality and efficiency by image processing, pattern recognition, machine learning, and computer vision methods~\cite{Ai-2021-ASRF}. Currently, the most widespread application of CAD is cancer global detection, which is implemented by image classification methods in computer vision~\cite{srinidhi2021deep}.

\begin{figure}[htbp!]
\centering
\includegraphics[trim={0cm 0cm 0cm 0cm},clip,width= 0.7\textwidth]{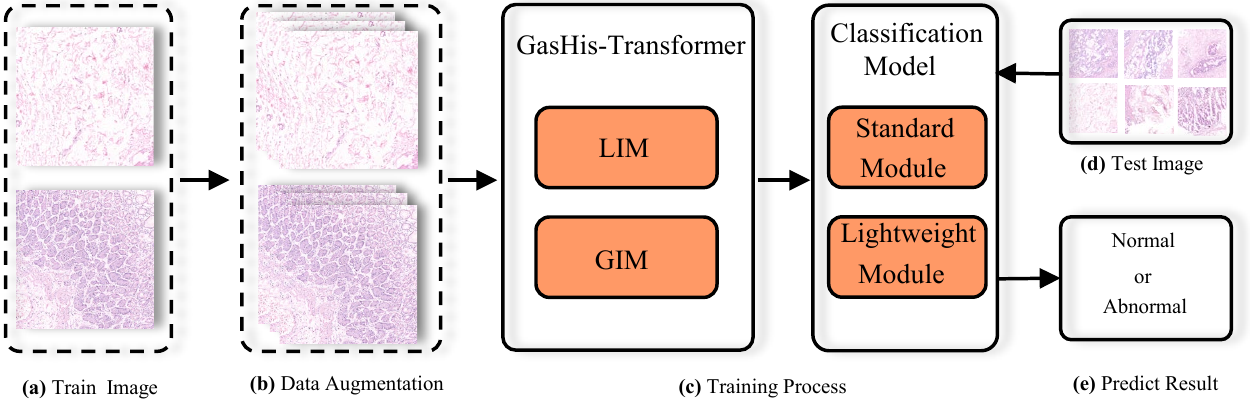}
\vspace{-0.3cm}
\caption{The architecture of the GasHis-Transformer model.}     
\label{fig:Architecture-GasHis}
\end{figure}

With the advent of artificial intelligence, deep learning has become the most extensive and widely used method for CAD~\cite{Ai-2021-ASRF}. Deep learning has been proved successful in many research fields, such as data mining, natural language processing and computer vision. It enables a computer to imitate human activities, solve complex pattern recognition problems and make excellent progress in artificial intelligence-related techniques. \emph{Convolutional Neural Network} (CNN) models are the dominant type of deep learning that can be applied to many computer vision tasks. However, there are some shortcomings of CNN models, one of which is that CNN models do not handle global information well. In contrast, the novel \emph{Visual Transformer} (VT) models applied in the field of computer vision can extract more abundant global information. In medicine, the composition of histopathological images are complex, with some abnormal images having a large portion of abnormal sections and some having a tiny portion of abnormal sections. Therefore, the model used for histopathological image global detection tasks must have a strong ability to extract global and local information. Considering the facts of CNN and VT models, a hybrid model has been heuristically proposed for \emph{Gastric Histopathological Image Detection} (GHID) tasks, namely GasHis-Transformer, to integrate the local and global information into an organic whole (Fig.~\ref{fig:Architecture-GasHis}).

The whole GasHis-Transformer model comprises two modules: Global Information Module (GIM) and Local Information Module (LIM). First, following the idea of BoTNet-50~\cite{srinivas2021bottleneck}, we have designed GIM to extract abundant global information to describe a gastric histopathological image as a whole. Then, the parallel structure idea of Inception-V3~\cite{szegedy2016rethinking} is followed to obtain multi-scale local information to represent the details of a gastric histopathological image.

\textbf{The contributions of this paper are as follows:} Firstly, considering the advantages of VT and CNN models, GasHis-Transformer model integrates the describing capability of global and local information of VT's and CNN's. Secondly, in GasHis-Transformer, the idea of multi-scale image analysis is introduced to describe the details of gastric histopathological images under a microscope. Furthermore, a lightweight module using the quantization method~\cite{choudhary2020comprehensive} and Dropconnect strategy~\cite{mobiny2021dropconnect} is heuristically proposed to reduce the model parameter size and training time for clinical applications with improved confidence. Finally, GasHis-Transformer not only obtains good global detection performance on gastric histopathological images but also shows an excellent generalization ability on histopathological image staging tasks for other cancers.

\vspace{-0.5cm}
\section{Related Work}
\label{section:rw}

\vspace{-0.3cm}
There have been many applications of GHID tasks in the field of pattern recognition. Traditional machine learning is an effective method that has been used for many years~\cite{Ai-2021-ASRF}. In the study of~\cite{sharma2017comparative}, random forest classifier is used to detect 332 global graph features including the mean, variance, skewness, kurtosis and other features extracted from gastric cancer histopathological images. In recent years, deep learning methods have become increasingly used in GHID tasks~\cite{Ai-2021-ASRF}. In the study of~\cite{WANG2019RMDL}, an improved ResNet-v2 network is used to detect images by adding an average pooling layer and a convolution layer. In the study of~\cite{song2020clinically}, 2166 whole slide images are detected by a deep learning model based on DeepLab-V3 with ResNet-50 architecture as the backbone.

Besides, there are many potential deep learning methods are possible for GHID tasks, such as AlexNet~\cite{krizhevsky2012imagenet}, VGG models~\cite{simonyan2014very}, Inception-V3 network~\cite{szegedy2015going}, ResNet models~\cite{he2016deep} and Xception network~\cite{chollet2017xception}. Especially, novel attention mechanisms show good global detection performance in image detection tasks, such as Non-local+Resnet~\cite{wang2018non}, CBAM+Resnet~\cite{woo2018cbam}, SENet+CNN~\cite{hu2018squeeze}, GCNet+Resnet~\cite{cao2019gcnet}, HCRF-AM~\cite{li2021hierarchical} and VT models~\cite{vaswani2017attention}. VT models are more and more used in image detection field~\cite{han2020survey}. There are two main forms of VT models in image detection tasks, that is the pure self-attention structure represented by Vision Transformer (ViT)~\cite{khan2021transformers} and the self-attention structure combined CNN models represented by BoTNet-50~\cite{srinivas2021bottleneck}, TransMed~\cite{dai2021transmed} and LeViT~\cite{graham2021levit}. The biggest advantage of VT models is that they perfectly solve the shortcomings of CNN models, where VT models can better describe the global information of images and have a good ability to extract global information by introducing an attention mechanism. 
\vspace{-0.5cm}
\section{GasHis-Transformer}
\label{section:mt}

\vspace{-0.3cm}
\subsection{Vision Transformer (ViT)}
\begin{figure}[htbp!]
\centering
\includegraphics[trim={0cm 0cm 0cm 0cm},clip,width=0.7\textwidth]{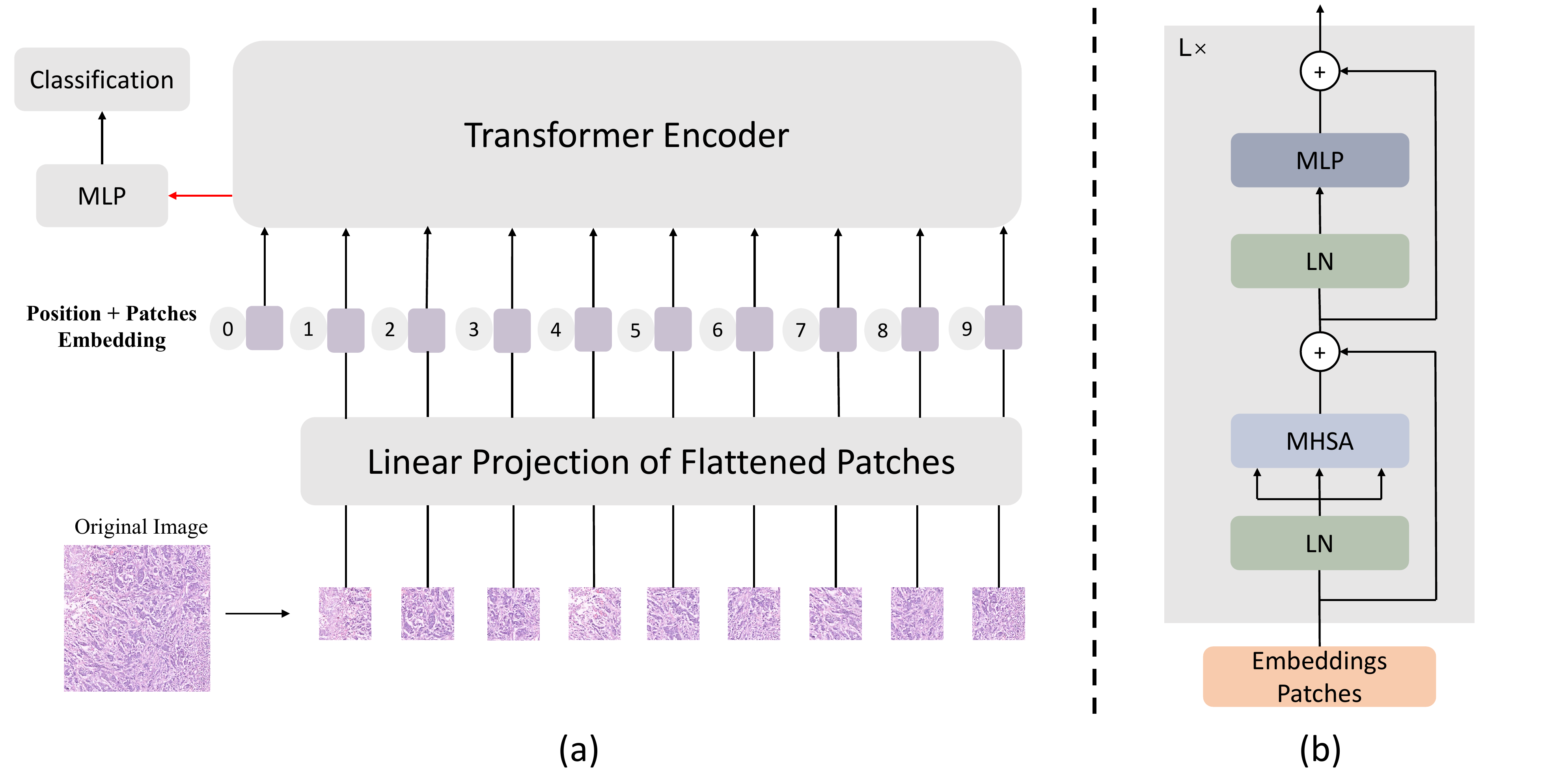}
\vspace{-0.5cm}
\caption{The details of ViT model. 
(a) is an overview of ViT model training process. 
(b) is a structure of transformer encoder. This architecture follows the idea of Fig. 1 in~\cite{dosovitskiy2020image}.}
\label{fig:Structure-ViT}
\end{figure}

The first model using a transformer encoder instead of standard convolution in the computer vision field is ViT~\cite{khan2021transformers,dosovitskiy2020image}. An overview of the ViT model is shown in Fig.~\ref{fig:Structure-ViT} (a). Image classification using ViT model can be divided into two stages: feature extraction stage and classification stage. In feature extraction stage, in order to handle a 2D image as a 1D sequence, 2D patch sequence $x_p\in \mathbb{R}^{N\times ({P^2\times C})}$ is obtained by reshaping the original image $x\in \mathbb{R}^{H\times W\times C}$. C is the number of image's channels, $(H\times W)$ is the size of each original image, $(P^2)$ is the size of each image patch, $N=HW/P^2$ is the sum of patch number as the same as input sequence length of the transformer encoder. The invariant hidden vector size $D$ used in the transformer goes through all layers, where all patches are flattened to $D$ dimensions, and $D$ dimensions (patch embeddings) are mapped by a linear projection that can be trained. To retain positional information, the sequence of embedding vectors combines standard 1D position embedding, and patch embeddings are selected to be the input of the transformer encoder.

The transformer encoder is composed of multiple alternative \emph{Multi Head Self-Attention} (MHSA) blocks~\cite{srinivas2021bottleneck} and multilayer perceptron (MLP) blocks~\cite{vaswani2017attention}. The structure of the transformer encoder is shown in Fig.~\ref{fig:Structure-ViT} (b). Layernorm (LN) is used in front of each layer and connected to the following block through residual connection. MLP block has two network layers connected by a non-linear Gaussian error linear units (GELU) activation function. Finally, in the classification stage, the output features after the feature extraction stage are passed through the fully connected layer composed of MLP to obtain the classification confidence.

\vspace{-0.3cm}
\subsection{BoTNet} 
\begin{figure}[htbp!]
\centering
\includegraphics[trim={0cm 0cm 0cm 0cm},clip,width=0.7\textwidth]{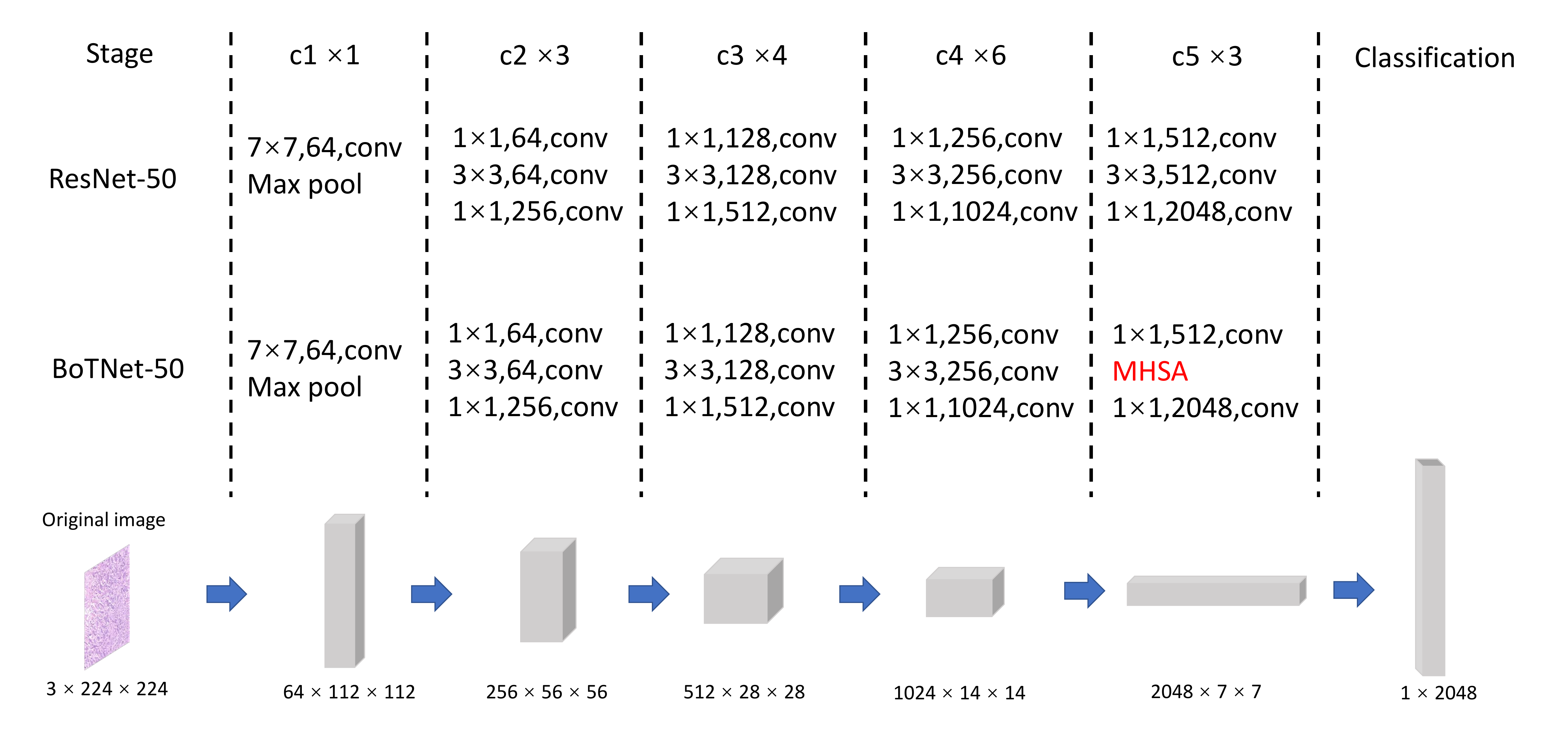}
\vspace{-0.5cm}
\caption{The architecture of BoTNet-50 and ResNet-50.}
\label{fig:Compare-BOTRES}
\end{figure}

BoTNet-50~\cite{srinivas2021bottleneck} is a VT model which combines ResNet-50 with the MHSA layer. Because the usage of the MHSA layers reduces massive parameters, BoTNet-50 is a network with a simple structure and powerful functions. The architecture of BoTNet-50 model compared to ResNet-50 is shown in Fig.~\ref{fig:Compare-BOTRES}. The process of the feature extraction blocks of BoTNet-50, which is the same as that of ResNet-50, is divided into five stages which are 1 set of stage c1, 3 sets of stage c2, 4 sets of stage c3, 6 sets of stage c4 and 3 sets of stage c5. Similarly to the hybrid model of ViT~\cite{dosovitskiy2020image}, in which input sequence extracted from CNN models to alter raw image patches, BoTNet-50 remains the model of ResNet-50 in advance of stage c4 and using the MHSA layers substitute for the last three $3 \times 3$ spatial convolutions in stage c5 of the model of ResNet. Thus, BoTNet-50 obtains the global self attention in 2D feature maps. The latter part is the same as ResNet-50. The average pooling layer and fully connected (FC) layer are used to extract features and obtain classification results.

\begin{figure}[ht]
\centering
\includegraphics[trim={0cm 0cm 0cm 0cm},clip,width=0.7\textwidth]{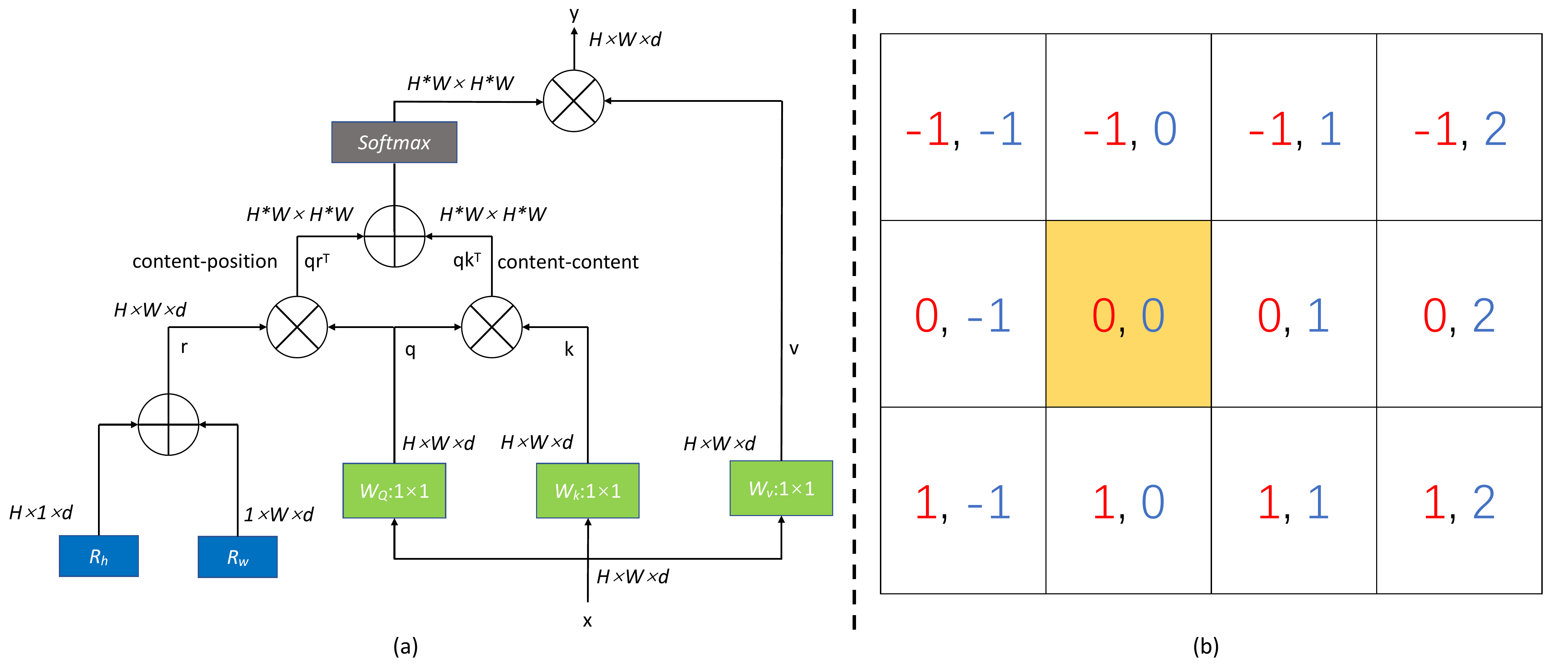}
\vspace{-0.5cm}
\caption{(a) is the structure of relative position encoding of the MHSA. 
$\oplus$ and $\otimes$ express sum and matrix multiply respectively. 
Blue blocks are location encoding. Green blocks are content encoding. 
(b) is is A single example of relative distance computation. 
The relative distance in the figure is calculated according to the bright position. Red is row offset and blue is column offset.}
\label{fig:MHSA}
\end{figure}

There are multiple differences between BoTNet-50 and ViT. The main difference is that the MHSA of ViT uses standard 2D patch sequence position encoding, while BoTNet-50 uses 2D relative position encoding. The latest results~\cite{bello2019attention} show that relative position encoding is more suitable for image classification tasks than traditional encoding. A structure of relative position encoding of the MHSA is shown in Fig.~\ref{fig:MHSA} (a). There are four sets of single-headed attention in each MHSA layer of BoTNet-50. At present, the structure of Fig.~\ref{fig:MHSA} (a) only takes one single-headed attention as example. First, for a given pixel $x_{i,j}\in \mathbb{R}$ , we extract $a,b\in N_k(i,j)$ from the spatial extent $k$ which centered on $x_{i,j}$. Second, $W_{Q}$, $W_{K}$ and $W_{V}$ are defined as the learnable transforms and can compute the queries $q_{i,j} = W_{Q}x_{i,j}$, keys $k_{a,b} = W_{K}x_{a,b}$ and values $v_{a,b} = W_{V}x_{a,b}$, which are linear transformations of the pixels of spatial extent. The content information multiply the queries and keys value vectors. Thirdly, $R_{h}$ and $R_{w}$ are defined as the separable relative position encodings of height and weight are expressed by row offset ($a-i$) and column offset ($b-j$). The row offset and column offset are shown in  Fig.~\ref{fig:MHSA} (b) and they are connected with an embedding $r_{a-i}$ and $r_{b-j}$. The row offset and column offset embeddings (position information) are connected to form $r_{a-i,b-j}$. Finally, the content information and position information are accumulated, and then the spatial-relative attention $y_{i,j}$ of the pixel $x_{i,j}$ is obtained by multiplying the aggregation results with values through softmax~\cite{shaw2018self} as shown in Eq.~\ref{eq:1}.
 \begin{equation}
  y_{i,j} = \sum_{a,b\in N_k(i,j)} softmax_{a,b}(q_{i,j}^\top k_{a,b}+ q_{i,j}^\top r_{a-i,b-j})v_{a,b}.
  \label{eq:1}
 \end{equation}

The number of parameters in the MHSA layer is different from that in the convolution layer. The number of parameters in convolution increases at a quadratic rate with the increase of spatial extent, while the MHSA layer does not change with the change of spatial extent. When the sizes of input and output are the same, the computational cost of the MHSA layer is far less than that of convolution in the same spatial extent. For example, when the input and output are 128-dimensional, the computational cost of 3 spatial extents in the convolution layer is the same as that of 19 spatial extents in the MHSA layer~\cite{shaw2018self}. Therefore, the parameters and computation time of BoTNet-50 is less than that of ResNet-50~\cite{srinivas2021bottleneck}.

\vspace{-0.3cm}
\subsection{GasHis-Transformer}
\begin{figure}[ht]
\centering
\includegraphics[trim={0cm 0cm 0cm 0cm},clip,width=0.7\textwidth]{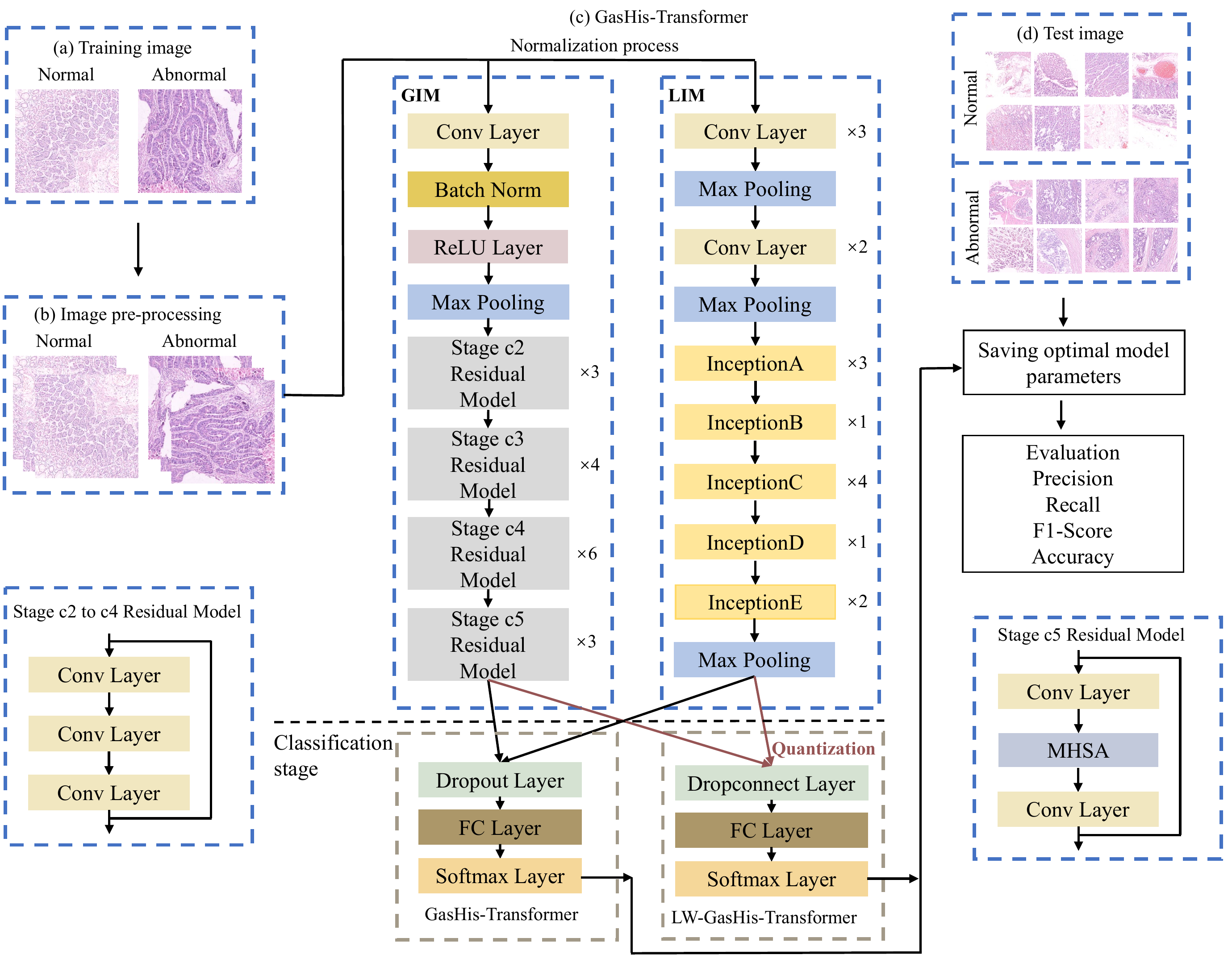}
\vspace{-0.5cm}
\caption{The structure of GasHis-Transformer model. (a) are training images including normal and abnormal. (b) is data pre-processing, which uses rotation and mirroring methods for data augmentation of training images. (c) are GasHis-Transformer and its lightweight version including the backbone network and the classification stage. (d) are test images including normal and abnormal.}
\label{fig:Structure-GasHis}
\end{figure}

\indent The GasHis-Transformer model and its lightweight version (LW-GasHis-Transformer) are proposed to detect gastric cancer in histopathological images, shown in Fig.~\ref{fig:Structure-GasHis}. The details of each block in LIM and GIM of GasHis-Transformer model are shown in Table~\ref{table:featuremap}. GasHis-Transformer applies image normalization to improve the image quality. This operation keeps the global information, only modifies the pixels to a specified range to accelerate the convergence of the training model.

\textbf{In Fig.~\ref{fig:Structure-GasHis}(a),} normal and abnormal gastric histopathological images are used as training data for GasHis-Transformer. 

\textbf{In Fig.~\ref{fig:Structure-GasHis}(b),} first, because of the multi-scale characteristics of histopathological images under the microscope, GasHis-Transformer augments the images by rotating and mirroring operations. Furthermore, GasHis-Transformer model applies image normalization to avoid this situation and speed up the model learning process and image normalization process is defined in Eq.~\ref{eq:2}.
\begin{equation}
  {\rm INPUT_{RGB}} = N({\rm IMG_{RGB}}),
  \label{eq:2}
\end{equation}
where ${\rm INPUT_{RGB}}$ and ${\rm IMG_{RGB}}$ represent original image and image which input into LIM and GIM, respectively. $N()$ is the image normalization processing. The shallow color of gastric cancer histopathological images and the implicit boundary characteristics of the nucleus results in poor image quality. This phenomenon is represented by the fact that any region of the whole image has a similar mean and standard deviation. The mean and standard deviation of a image are defined in Eq.~\ref{eq:3} and Eq.~\ref{eq:4}.
\begin{equation}
 \mu({\rm IMG_{RGB}}) = \frac{\sum_{i=1,j=1}^{N}{\rm IMG_{RGB}}(i,j)}{N \times N}, 
  \label{eq:3}
\end{equation}
 \begin{equation}
  \sigma({\rm IMG_{RGB}}) = \frac{\sqrt{\sum_{i=1,j=1}^{N}[{\rm IMG_{RGB}}(i,j)-\mu({\rm IMG_{RGB}})]^{2}}}{N \times N}, 
  \label{eq:4}
\end{equation}
where ${\rm IMG_{RGB}}$ is an original image with $N \times N$ sizes and ${\rm IMG_{RGB}}(i,j)$ is the pixel of this image, $\mu({\rm IMG_{RGB}})$ and $\sigma({\rm IMG_{RGB}})$ represent the mean and standard deviation of this image, respectively. According to the theory of convex optimization and data probability distribution, the image is operated according to Eq.~\ref{eq:5} to finally obtain a normalized image with a mean of 0 and a standard deviation of 1.
\begin{equation}
  {\rm INPUT_{RGB}} = \frac{{\rm IMG_{RGB}} - \mu({\rm IMG_{RGB}})}{\sigma({\rm IMG_{RGB}})}.
  \label{eq:5}
\end{equation}
When the input image pixels of all samples are positive, the weights from the same convolution kernel can only increase or decrease simultaneously, and ReLU layer can be shielded negative weights from convolution layer, resulting in a slow learning speed. Each pixel in images using image normalization is related to the global mean and standard deviation, preserving the image's global information and nonlinear features. It enables GasHis-Transformer model to detect the region of interest faster when training the model, thus improving convergence speed and detection accuracy of the model. 

\textbf{In Fig.~\ref{fig:Structure-GasHis}(c),} images are used to train the proposed model, and this step is the core of the whole structure. GasHis-Transformer includes two parts: Global Information Module (GIM) and Local Information Module (LIM). In GIM, GasHis-Transformer follows the idea of BoTNet-50 such that convolution layer in the last residual layer of the ResNet-50 model is replaced by the MHSA~\cite{srinivas2021bottleneck}. GIM retains all the structures before c5 stage of BoTNet-50 and 2048-dimensional global features are extracted in the last pooling layer of GIM. In LIM, GasHis-Transformer follows the idea of Inception-V3 and carries out a series of modifications to the traditional model. To match the standard input of GIM and make the features extracted by GIM and LIM in the whole network with the same measurement, LIM modifies the standard input size of Inception-V3~\cite{szegedy2016rethinking} model from $299 \times 299$ to $224 \times 224$ and modifies the standard output size of every convolution layer and pooling layer in GasHis-Transformer. Similar to GIM, 2048-dimensional local features are extracted in the last pooling layer of LIM. Compared to GasHis-Transformer, 32-bit floating-point GasHis-Transformer parameters are degraded to 16-bit floating-point numbers via quantization in the training stage of LW-GasHis-Transformer~\cite{choudhary2020comprehensive}. At the end of GIM and LIM, the global and local features are fused to obtain the 4096-dimensional splicing feature as the final feature trained. 

In the classification stage, an optimization layer is added to suppress the risk of overfitting and retain the global and local information before the FC layer. The optimization layer optimizes the model during the training and testing stages: (1) In GasHis-Transformer, the optimization layer uses Dropout. In the testing process, the model with Dropout can be considered as performing simultaneous prediction using multiple classification networks with shared parameters, which can significantly improve the generalizability of the classification task~\cite{srivastava2014dropout}. (2) In LW-GasHis-Transformer, the optimization layer uses Dropconnect, which is a generalization of Dropout for regularizing neural networks. Dropconnect replaces Dropout for approximate Bayesian inference being more capable of extracting uncertainty~\cite{lyu2022softdropconnect}. Dropconnect discards the weights between hidden layers according to a fixed probability instead of simply discarding hidden nodes and samples the weights of each node with a Gaussian distribution in the testing stage~\cite{mobiny2021dropconnect}. Dropconnect can effectively solve problems caused by model quantization. Finally, the fused features go through the FC and Softmax layers to obtain the final classification confidence.

\textbf{In Fig.~\ref{fig:Structure-GasHis}(d),} test images including normal and abnormal are used to evaluate the global detection performance of GasHis-Transformer. 
\begin{table}[htbp]
		\tiny
		\centering
		\caption{The details of each block in GIM and LIM of GasHis-Transformer model.}
		\label{table:featuremap}
		\setlength{\tabcolsep}{2mm}
		\vspace{-0.3cm}
		{
		\begin{tabular}{|lll|lll|}
\hline
\multicolumn{3}{|l|}{GIM}                                                                                      & \multicolumn{3}{l|}{LIM}                                                  \\ \hline
\multicolumn{2}{|l|}{Block}                                      & Output Feature                              & \multicolumn{2}{l|}{Block}                   & Output Feature             \\ \hline
\multicolumn{2}{|l|}{Conv(7,7)}                                  & $112 \times 112 \times 64$                  & \multicolumn{2}{l|}{Conv(3,3)}               & $111 \times 111 \times 32$ \\ \hline
\multicolumn{2}{|l|}{Batch Norm}                                 & $112 \times 112 \times 64$                  & \multicolumn{2}{l|}{Conv(3,3)}               & $109 \times 109 \times 32$ \\ \hline
\multicolumn{2}{|l|}{ReLU Layer}                                 & $112 \times 112 \times 64$                  & \multicolumn{2}{l|}{Conv(3,3)}               & $109 \times 109 \times 64$ \\ \hline
\multicolumn{2}{|l|}{Max Pooling}                                & $56 \times 56 \times 64$                    & \multicolumn{2}{l|}{Max Pooling}             & $54 \times 54 \times 64$   \\ \hline
Stage c2       & \multicolumn{1}{l|}{\multirow{2}{*}{$\times$3}} & \multirow{2}{*}{$56 \times 56 \times 256$}  & \multicolumn{2}{l|}{Conv(1,1)}               & $54 \times 54 \times 80$   \\ \cline{4-6} 
Residual Model & \multicolumn{1}{l|}{}                           &                                             & \multicolumn{2}{l|}{Conv(3,3)}               & $52 \times 52 \times 192$  \\ \hline
Stage c3       & \multicolumn{1}{l|}{\multirow{2}{*}{$\times$4}} & \multirow{2}{*}{$28 \times 28 \times 512$}  & \multicolumn{2}{l|}{Max Pooling}             & $25 \times 25 \times 192$  \\ \cline{4-6} 
Residual Model & \multicolumn{1}{l|}{}                           &                                             & Inception A & \multicolumn{1}{l|}{$\times$3} & $25 \times 25 \times 256$  \\ \hline
Stage c4       & \multicolumn{1}{l|}{\multirow{2}{*}{$\times$6}} & \multirow{2}{*}{$14 \times 14 \times 1024$} & Inception B & \multicolumn{1}{l|}{$\times$1} & $25 \times 25 \times 288$  \\ \cline{4-6} 
Residual Model & \multicolumn{1}{l|}{}                           &                                             & Inception C & \multicolumn{1}{l|}{$\times$4} & $12 \times 12 \times 768$  \\ \hline
Stage c5       & \multicolumn{1}{l|}{\multirow{2}{*}{$\times$3}} & \multirow{2}{*}{$7 \times 7 \times 2048$}   & Inception D & \multicolumn{1}{l|}{$\times$1} & $5 \times 5 \times 1024$   \\ \cline{4-6} 
Residual Model & \multicolumn{1}{l|}{}                           &                                             & Inception E & \multicolumn{1}{l|}{$\times$2} & $5 \times 5 \times 2048$   \\ \hline
\end{tabular}
		}
\end{table}

\vspace{-0.7cm}
\section{Experiment Results and Analysis}\label{section:ea}

\vspace{-0.3cm}
\subsection{Experimental Settings}

\vspace{-0.2cm}
\subsubsection{Dataset}
\begin{figure}[ht]
\centering
\includegraphics[trim={0cm 0cm 0cm 0cm},clip,width= 0.7\textwidth]{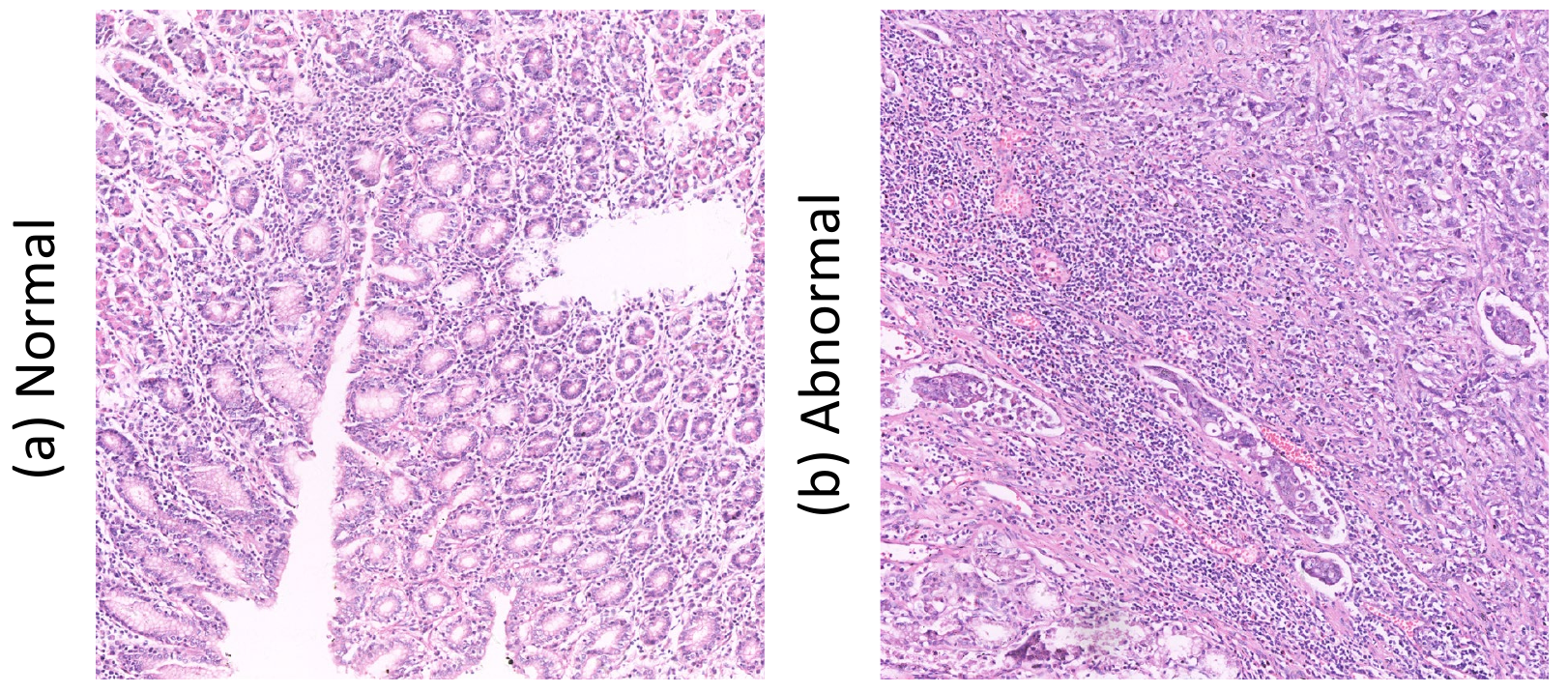}
\caption{Normal and abnormal examples in the HE-GHI-DS.}
\vspace{-0.5cm}
\label{fig:Example-Gastric}
\end{figure}

In this paper, an open-source Hematoxylin and Eosin (H\&E) stained gastric histopathological image dataset(HE-GHI-DS) is used in the experiment to evaluate the global detection performance of GasHis-Transformer$\footnote{This dataset is open access on: Sun, C. and Li, C. and Li, Y, Data for hcrf, https://data.mendeley.com/datasets/thgf23xgy7/2}$. The hematoxylin-stained solution is alkaline, which makes chromatin in the nucleus and ribosome in cytoplasm purple-blue; Eosin-stained solution is acidic, which makes the components in the cytoplasm and extracellular matrix red. The images are part of the whole slide images in `*.tiff' format, where they are magnified 20 times and the image size is $2048 \times 2048$ pixels~\cite{li2018deep}. HE-GHI-DS includes 140 normal images and 560 abnormal images. Some examples of normal and abnormal gastric histopathological images are shown in Fig.~\ref{fig:Example-Gastric}. In the normal images, the nuclei are stable and arranged regularly and the nucleo-cytoplasmic ratio is small~\cite{Ai-2021-ASRF}. On the contrary, in the abnormal images, the nucleus is abnormally large and irregular in features of dish or crater.

\vspace{-0.3cm}
\subsubsection{Data Settings}
Due to the imbalance of the initial training data in HE-GHI-DS, deep learning models only learn the characteristics of one category, leading to low classification accuracy and weak generalization ability of models~\cite{kim2021normalized}. In order to balance the training images, 140 abnormal images are randomly selected from all 560 abnormal images to match the number of normal images. In the GHID task, GasHis-Transformer equally uses 140 abnormal images and 140 normal images. Moreover, the abnormal and normal images in the dataset are randomly partitioned into training, validation and test sets with a ratio of 1: 1: 2. Furthermore, all images are flipped horizontally and vertically and rotated 90, 180, 270 degrees to augment the training, validation and test datasets to six times. In addition, although some information is lost by direct image resize operations, it shows that deep learning networks are robust to different sizes of pathological images in our previous study~\cite{liu2021ITARO}. Therefore, all images are resized into $224 \times 224$ pixels by bilinear interpolation. Unlike some other GHID tasks, because rotation and mirroring operations change the relative positions of cancers in histopathological images, both the validation and test sets are expanded to verify the multi-scale generalization ability of the GasHis-Transformer. The data settings and augmentation results are shown in Table~\ref{table:1}.
\begin{table}[htbp!]
		\tiny
		\centering
		\caption{Data setting for training, validation and test sets.}
		\vspace{-0.3cm}
	\label{table:1}
	\begin{tabular}{|c|c|c|c|c|c|}
	\hline
	\multicolumn{2}{|c|}{Image Type}      & Training & Validation & Test & Sum \\ \hline
	\multirow{2}{*}{Normal}   & Oringin   & 35    & 35         & 70   & 140 \\ \cline{2-6} 
	                          & Augmented & 210   & 210        & 420  & 840 \\ \hline
	\multirow{2}{*}{Abnormal} & Oringin   & 35    & 35         & 70   & 140 \\ \cline{2-6} 
	                          & Augmented & 210   & 210        & 420  & 840 \\ \hline
	\end{tabular}
\end{table}

\vspace{-0.5cm}
\subsubsection{Hyper-parameter Setting}
\label{HPS}
GasHis-transformer and LW-GasHis-Transformer are used to train the gastric histopathological image dataset for 75 epochs. In each epoch, batch size is set to 16. It uses an approach to train from scratch for the GHID task. AdamW optimizer is used for optimization and its parameters are set as: $2e-3$ learning rate, $1e-8$ eps, $[0.9,0.999]$ betas and $1e-2$ weight decay. Especially, VGGNets are trained at the learning rate of $2e-4$. AdamW solves the problem of parameter over-fitting with Adam optimizer by introducing L2 regularization terms of parameters in the loss function. It is the fastest optimizer for gradient descent speed and training neural networks which are used in all models. A learning rate adjustment strategy is used, that is, if the set loss function is not decreased within 15 epochs, the learning rate is reduced by ten times. In addition, the ratio of Dropout and Dropconnect are both set to 0.5 in the training process.

\vspace{-0.2cm}
\subsubsection{Evaluation Criteria}
Precision (Pre), recall (Rec), F1-score (F1) and accuracy (Acc) are used to evaluate the GasHis-Transformer model, where true positive (TP), true negative (TN), false positive (FP) and false negative (FN) are used in the definition of these four criteria in Table~\ref{table:3}. 
\begin{table}[ht]
		\tiny
		\centering
		\renewcommand\arraystretch{1.5}
		\caption{Criteria and corresponding definitions for image global detection evaluation.}
	\label{table:3}
	\vspace{-0.3cm}
	\begin{tabular}{|c|c|c|c|}
		\hline
		Criterion & Definition & Criterion & Definition \\\hline
		Pre & $\rm \frac{TP}{TP+FP}$ & Rec & $\rm \frac{TP}{TP+FN}$ \\\hline
		F1 & $\rm \frac{2\times TP}{2\times TP+FP+FN}$ & Acc & $\rm \frac{TP+TN}{TP+TN+FP+FP}$\\\hline
	\end{tabular} 
\end{table}

\vspace{-0.5cm}
\subsection{Evaluation Results of GasHis-Transformer}

\vspace{-0.2cm}
\subsubsection{Experimental Results}
\label{subsubsection:er}
\begin{figure}[ht]
\centering
\includegraphics[trim={0cm 0cm 0cm 0cm},clip,width= 0.7\textwidth]{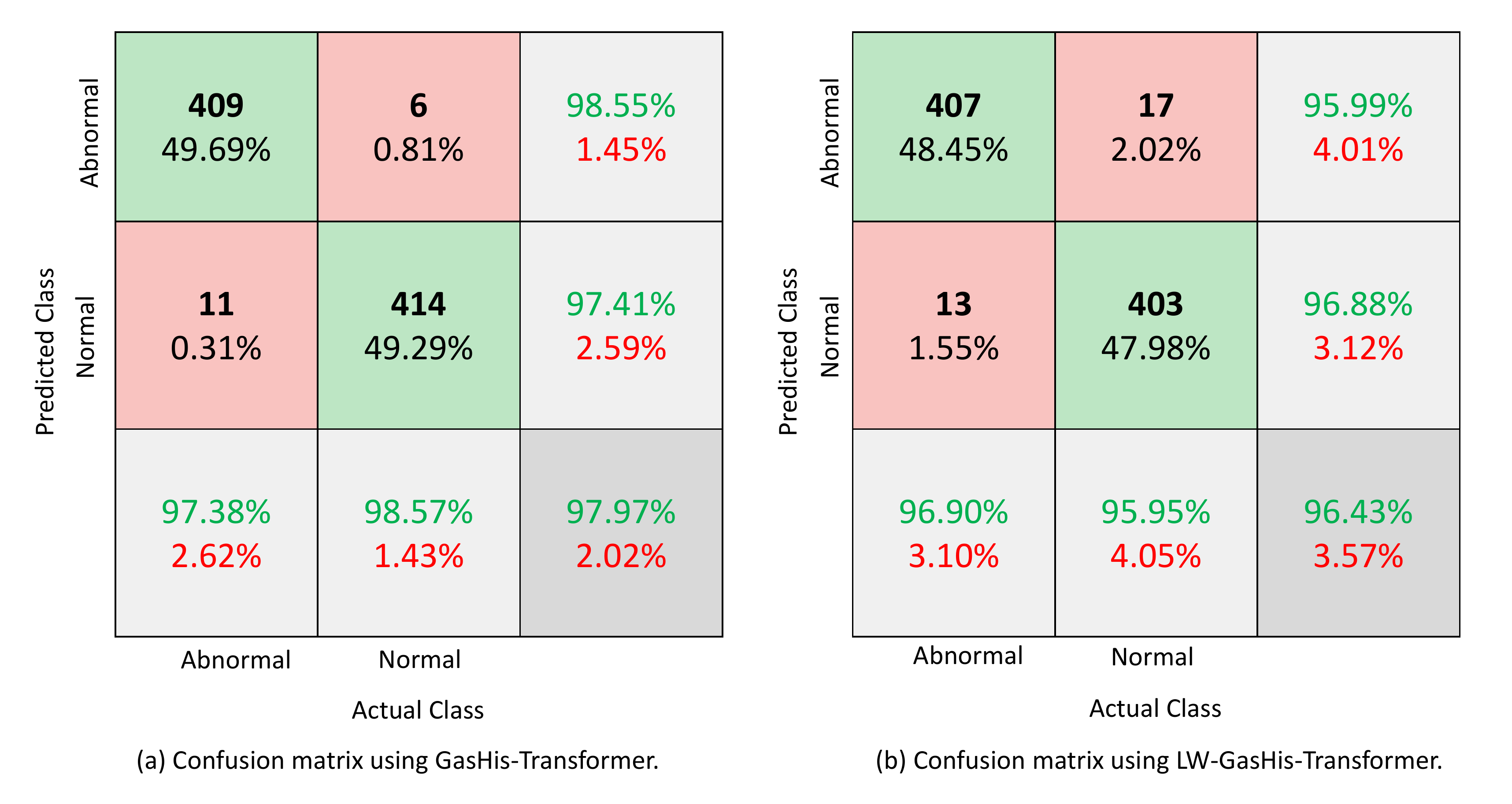}
\vspace{-0.5cm}
\caption{Confusion matrix using GasHis-Transformer and LW-GasHis-Transformer,  respectively. 
Green and red numbers are the percentage of correct and incorrect cases, respectively.}
\label{fig:Confusionmatrix}
\end{figure}

The criteria of the GasHis-Transformer and LW-GasHis-Transformer are calculated respectively to determine whether the models converge and have generalize well. The average confusion matrix of five randomized experiments on GasHis-Transformer and LW-GasHis-Transformer is shown in Fig.~\ref{fig:Confusionmatrix}. In Fig.~\ref{fig:Confusionmatrix}(a), 409 abnormal images and 414 normal images are correctly classified into the correct categories. Only 11 abnormal images are incorrectly reported as normal, and 6 normal image are incorrectly detected as abnormal. Overall, Pre, Rec, F1 and Acc of the global detection using GasHis-Transformer on the test set are 98.55\%, 97.38\%, 97.97\% and 97.97\%, respectively. In Fig.~\ref{fig:Confusionmatrix}(b), 407 abnormal images and 403 normal images are correctly classified into the correct categories. Only 13 abnormal images are incorrectly reported as normal, and 17 normal images are incorrectly detected as abnormal. Pre, Rec, F1 and Acc of the global detection using LW-GasHis-Transformer are 95.99\%, 96.90\%, 96.43\% and 96.43\%, respectively.

\begin{figure}[htbp]
\centering
\includegraphics[trim={0cm 0cm 0cm 0cm},clip,width= 0.8\textwidth]{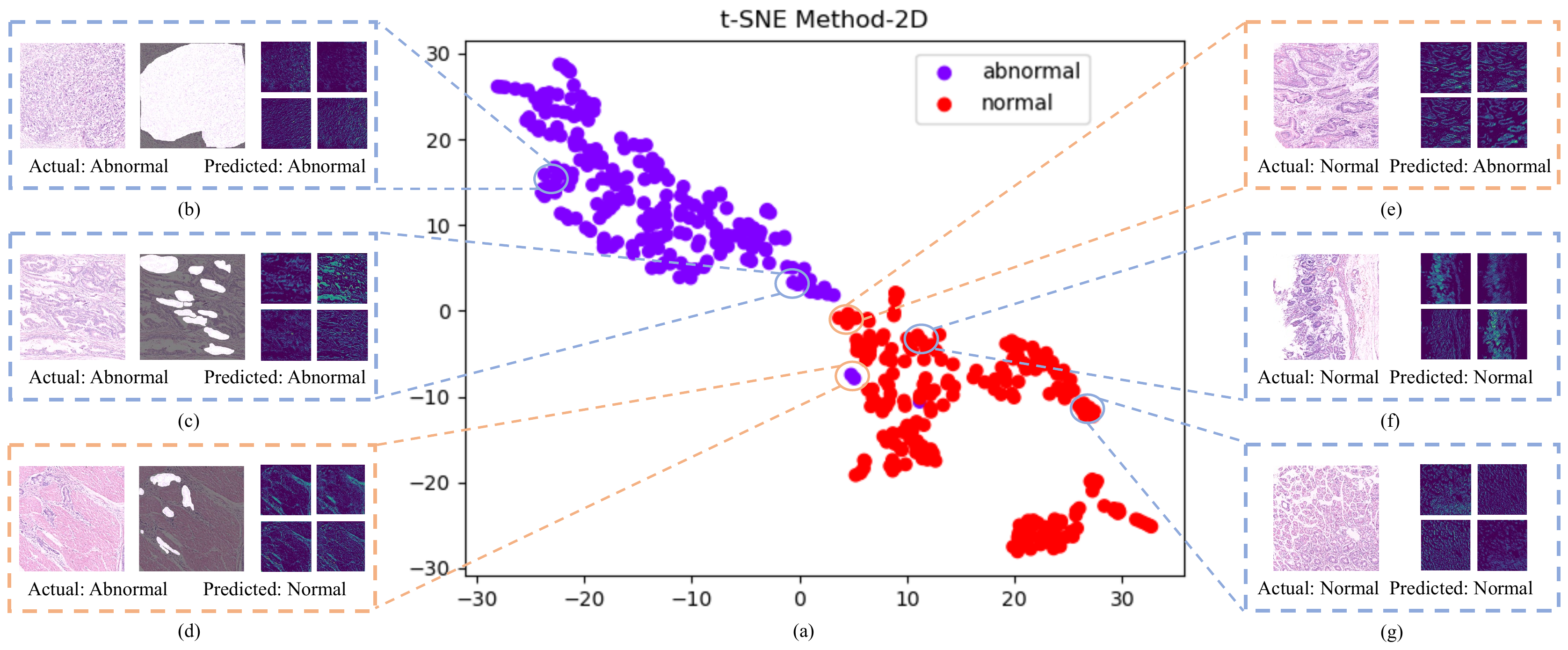}
\vspace{-0.5cm}
\caption{Visualization analysis of the misidentification results. 
(a) is 2-D vector scatter plots using the t-SNE method. 
(b)-(d) are abnormal images. Left column are original images, middle column are pixel-level ground 
truth images and right column are four feature maps of each image. 
(e)-(g) are normal images. 
Left column are original images and right column are four feature maps of each image. 
Light blue color indicates correctly detected images, orange color indicates incorrectly 
detected images.}
\label{fig:Example-Mis}
\end{figure}

In order to explain the performance of GasHis-Transformer in the GHID task and to analyze the causes of misidentification, we compress the 4096-dimensional feature vectors of each image to a 2-dimensional (2-D) space using the t-SNE method for analysis of the first randomized experiment~\cite{van2008visualizing}. The 2-D vector scatter plots obtained using t-SNE method are shown in Fig.~\ref{fig:Example-Mis}(a), while Fig.~\ref{fig:Example-Mis}(b)-(g) show the images represented by different positions in the scatter plots and their feature maps, respectively.

First, features extracted by GasHis-Transformer model can distinguish most abnormal and normal images as shown in Fig.~\ref{fig:Example-Mis}(a). For example, in Fig.~\ref{fig:Example-Mis}(b) and Fig.~\ref{fig:Example-Mis}(g), they are taken from clusters of abnormal and normal images, respectively. Most abnormal images have extensive carcinoma areas and are highly differentiated without prominent lumen structures, just like Fig.~\ref{fig:Example-Mis}(b). Most normal images have the lumen structures excluding the interstitium and background, just like Fig.~\ref{fig:Example-Mis}(g). Meanwhile, the feature maps in Fig.~\ref{fig:Example-Mis}(b) and Fig.~\ref{fig:Example-Mis}(g) show that the weights of GasHis-Transformer tend to favor cancerous regions and lumen structures for most abnormal and normal images, respectively. It demonstrates the effectiveness of GasHis-Transformer model in the GHID task.

Additionally, there are a small number of abnormal and normal images with similar 2-D feature vectors, which is the cause of misidentification by the GasHis-Transformer model. For example, as in Fig.~\ref{fig:Example-Mis}(c)-(f), these four images have 2-D similar feature vectors. Fig.~\ref{fig:Example-Mis}(c) only contains a small portion of the cancerous regions. The feature maps clearly show that the model detects many background regions but still accurately detects the cancerous regions, which indicates that the GasHis-Transformer model is robust to background information and prefers to detect large contiguous regions. Fig.~\ref{fig:Example-Mis}(d) has a smaller cancerous region than Fig.~\ref{fig:Example-Mis}(c). The feature map shows that the GasHis-Transformer model, which detects large connected regions, has difficulty detecting the tiny cancerous regions, leading to the final identification mistake. Fig.~\ref{fig:Example-Mis}(e) shows intestinal epithelial metaplasia, which is the last normal staging before cancerous. Fig.~\ref{fig:Example-Mis}(e) and Fig.~\ref{fig:Example-Mis}(d)-(c) are already very similar in visual perspective, so the GasHis-Transformer model misidentified this image. The feature map in Fig.~\ref{fig:Example-Mis}(f) shows that although interstitium is detected in the feature map, the model assigns more weight to the lumen structure, indicating that GasHis-Transformer model is robust not only to the background information but also to the interstitial information.

\vspace{-0.5cm}
\subsubsection{Contrast Experiment of GHID}
In order to show the effectiveness of GasHis-Transformer and LW-GasHis-Transformer in the GHID task, a series of comparative experiments are carried out using CNNs and attention models on the testing set. In addition, classical CNNs and attention models are compared with and without image normalization, where all hyper-parameters are set to the same values as that in Sub-section~\ref{HPS}.
\paragraph{\textbf{Comparison with Other Models:}}
The GHID results of all models are compared in Table~\ref{table:4}. Frist, it is obvious that GasHis-Transformer achieves good performance in terms of Pre, F1 and Acc. GasHis-Transformer has some improvement compared to Xception, which shows good performance compared to other traditional CNN models. GasHis-Transformer has a higher Rec, F1, and Acc than the Xception model. Second, although the performance of LW-GasHis-Transformer is degraded compared with that of GasHis-Transformer, it still is better than CNN and attention models. In addition, since GasHis-Transformer and LW-GasHis-Transformer can extract multi-scale features, they have more stable training results and minor variance than CNN and attention mechanism models that extract features on the same scale. Finally, although Transformer models using sequential CNN such as BotNet-50~\cite{srinivas2021bottleneck}, TransMed~\cite{dai2021transmed} and LeViT~\cite{graham2021levit} have higher results than pure Transformer models, they lose more information than GasHis-Transformer model. Therefore, Transformer models using sequential CNN are less effective than GasHis-Transformer model. 
\begin{table}[htbp]  
		\tiny
		\centering
		\caption{A comparison of different models on the HE-GHI-DS test set. ([In \%].)}
		\label{table:4}
		\vspace{-0.3cm}
		{
		\begin{tabular}{|c|c|c|c|c|}
			\hline
			Models & Pre & Rec & F1 & Acc\\
			\hline
			GasHis-Transformer & $\bm{98.55\pm1.07}$ & $97.38\pm1.33$ & $\bm{97.97\pm0.78}$ & $\bm{97.97\pm0.74}$ \\\hline
			LW-GasHis-Transformer & $95.99\pm2.64$ & $96.90\pm2.96$ & $96.43\pm0.98$ & $96.43\pm1.39$ \\\hline
			Xception~\cite{chollet2017xception} & $94.48\pm3.21$ & $\bm{97.78\pm2.25}$ & $95.98\pm1.31$ & $95.94\pm1.36$ \\\hline
			ResNet-50~\cite{he2016deep} & $93.40\pm2.44$ & $95.26\pm1.94$ & $94.26\pm1.43$ & $94.24\pm1.43$\\\hline
			Inception-V3~\cite{szegedy2016rethinking} & $93.64\pm2.80$ & $94.40\pm3.83$ & $93.96\pm0.51$ & $93.80\pm0.54$\\\hline
			VGG-16~\cite{simonyan2014very} & $90.82\pm3.73$ & $94.48\pm3.86$ & $92.38\pm2.79$ & $92.34\pm2.79$\\\hline
			VGG-19~\cite{simonyan2014very} & $88.68\pm2.66$ & $94.68\pm3.32$ & $91.34\pm2.10$ & $91.24\pm2.25$\\\hline
			ViT~\cite{dosovitskiy2020image} & $86.10\pm3.88$ & $83.72\pm6.09$ & $84.88\pm1.24$ & $84.78\pm1.27$\\\hline
			BotNet-50~\cite{srinivas2021bottleneck} & $87.72\pm2.29$ & $90.56\pm2.88$ & $88.84\pm0.60$ & $88.88\pm0.64$ \\\hline
			TransMed~\cite{dai2021transmed} & $94.34\pm2.06$ & $97.06\pm2.27$ & $95.58\pm0.64$ & $95.58\pm0.64$\\\hline
			LeViT~\cite{graham2021levit} & $91.90\pm1.28$ & $90.50\pm3.10$ & $91.26\pm1.63$ & $91.26\pm1.60$\\\hline
			HCRF-AM~\cite{li2021hierarchical} & $92.90\pm2.51$ & $91.94\pm8.26$ & $92.06\pm5.50$ & $94.24\pm1.83$ \\\hline
			GCNet+Resnet~\cite{cao2019gcnet} & $96.82\pm2.64$ & $96.40\pm2.89$ & $95.26\pm1.19$ & $96.48\pm0.17$ \\\hline
			SENet+CNN~\cite{hu2018squeeze} & $95.94\pm1.36$ & $95.94\pm1.36$ & $95.94\pm1.36$ & $95.94\pm1.36$ \\\hline
			CBAM+Resnet~\cite{woo2018cbam} & $94.22\pm2.83$ & $96.10\pm2.91$ & $94.00\pm3.06$ & $95.04\pm1.91$ \\ \hline
			Non-local+Resnet~\cite{wang2018non} & $94.46\pm2.63$ & $97.00\pm2.78$ & $94.20\pm2.75$ & $95.58\pm1.10$ \\\hline
		\end{tabular}
		}
\end{table}

\begin{figure}[ht]
\centering
\includegraphics[trim={0cm 0cm 0cm 0cm},clip,width= 0.65\textwidth]{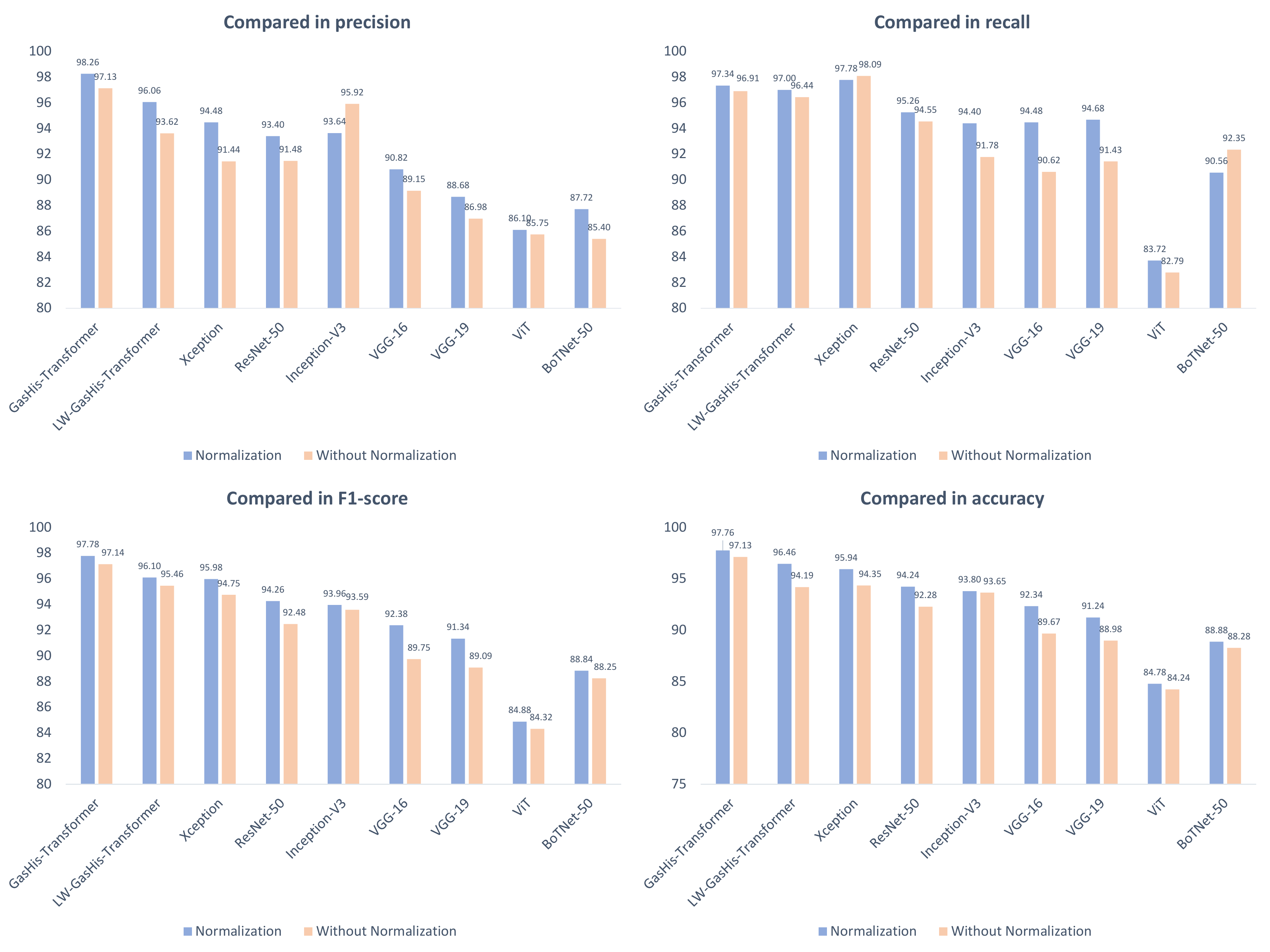}
\vspace{-0.5cm}
\caption{Effect of normalization in during preprocessing raw data. 
All images in training, validation and test sets are normalized according to 
Eq.~\ref{eq:5}.}
\label{fig:Compare-preprocess}
\end{figure}

\vspace{-0.5cm}
\paragraph{\textbf{Effect of Normalization on Model Performance:}}
The comparison results using normalization during pre-processing is shown in Fig.~\ref{fig:Compare-preprocess}. There is an increasing trend of Pre, Rec, F1 score, and Acc when using normalization. For Pre, all models have improvement except Inception-V3, and the proposed GasHis-Transformer and LW-GasHis-Transformer have improved by 1.11\% and 2.46\%, respectively. For Rec, all models have better results except Xception and BoTNet-50, and GasHis-Transformer and LW-GasHis-Transformer have improved by 0.44\% and 0.55\%, respectively. In summary, normalization in image preprocessing can improve detection performance.

\vspace{-0.5cm}
\subsubsection{Robustness Test of GasHis-Transformer}
\begin{figure}[ht]
\centering
\includegraphics[trim={0cm 0cm 0cm 0cm},clip,width= 0.65\textwidth]{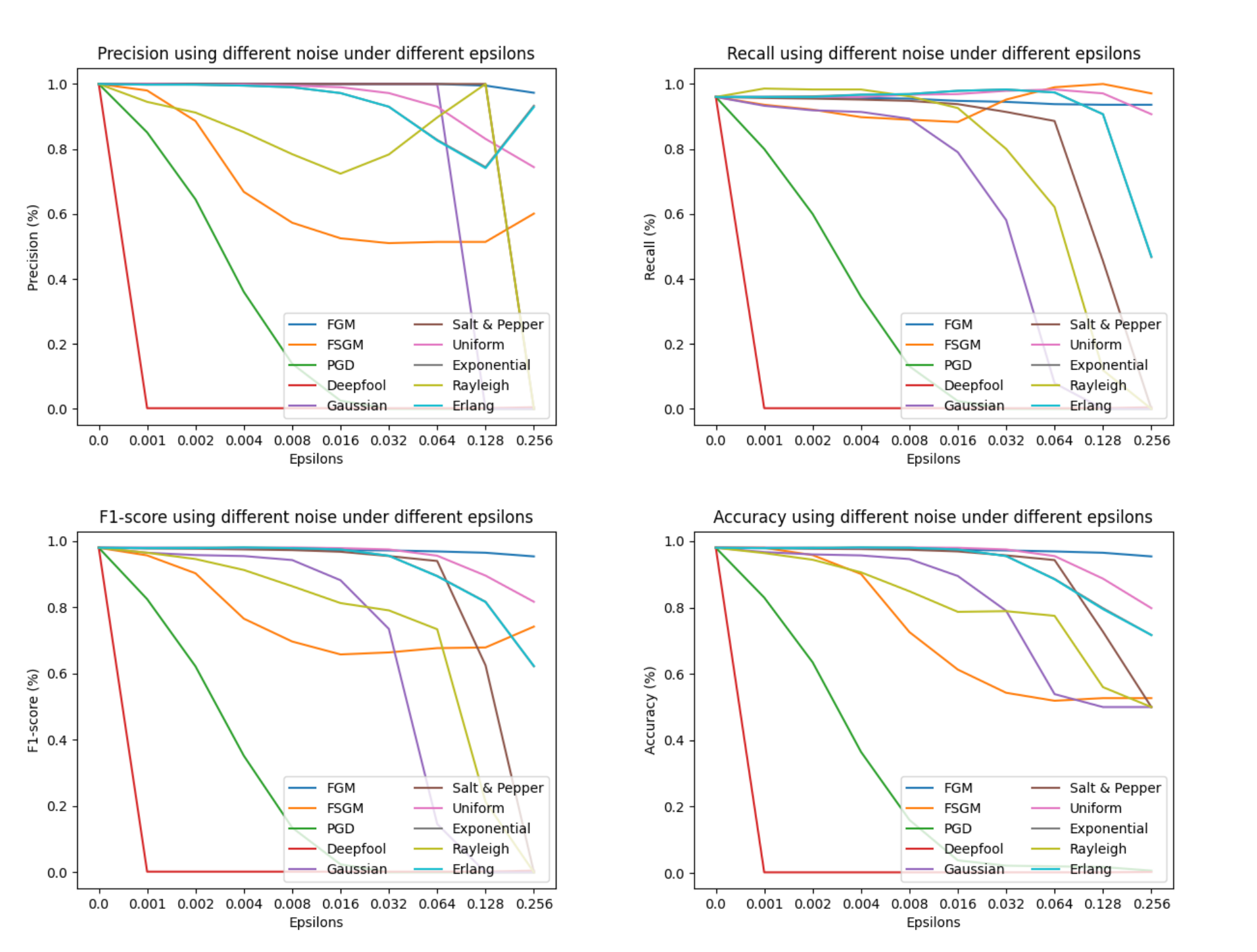}
\vspace{-0.5cm}
\caption{Robustness test of GasHis-Transformer under ten adversarial attack and conventional noises.}
\label{fig:Noise}
\end{figure}

Robustness is a property that maintains the stability of a model under parameter ingestion and 
measures the behavior of systems under non-standard conditions. Robustness is defined by 
community as the degree to which a system operates correctly in the presence of exceptional 
inputs or stressful environmental conditions. The robustness test aims to work correctly with 
each functional module when handling incorrect data and abnormal problems (through adding noise 
or taking other datasets), enhancing models’ fault resistance. To test the robustness of the 
proposed GasHis-Transformer model, ten different adversarial attacks and conventional noises 
are added to the HE-GHI-DS test set. Adversarial attacks are subtle interference added to 
the input sample that causes CNN model to give an incorrect output with a high confidence 
level~\cite{ghosh2022black}. Adversarial attacks include FGM~\cite{goodfellow2014EAHAE}, 
FSGM~\cite{kurakin2016adversarial}, PGD~\cite{madry2017TDLMR} and Deepfool~\cite{moosavi2016DASAA}; conventional noises include Gaussian, Salt \& Pepper, uniform, exponential, Rayleigh and 
Erlang noise. First, the epsilons are performed with nine levels in $[0.001,0.256]$ using 
0.001 as initialization and the powers of 2 as step length. Then, Pre, Rec, F1 and Acc are 
used to evaluate the robustness of GasHis-Tranformer in the GHID task. Fig.~\ref{fig:Noise} 
shows four criteria under different epsilons and noise.

For adversarial attacks noises, first, GasHis-Transformer is optimally robust when FGM is increased, 
and different epsilons have almost no effect on the model. Secondly, although criteria obtained 
by adding FSGM have some differences compared with adding FGM, in general, the performance is 
positive. When epsilon is higher than 0.032, the criteria converge to stability, even Rec 
increases slightly. Finally, adding Deepfool and PGD of any magnitude of epsilons results in 
a poor detection of the model. In summary, for noise generation by adversarial attacks, 
GasHis-Transformer has better robustness to FGM and FSGM.

For conventional noises, first, while four criteria of adding Erlang noise and uniform noise 
decrease compared to those of FGM in adversarial attacks, they are relatively constant compared 
to those with other convention noises. Therefore, adding them does not affect the robustness 
of the model in general. In addition, when epsilon is lower than 0.1, the model is barely 
affected by adding Gaussian, Rayleigh and Salt \& Pepper noise. However, when epsilon is higher 
than 0.1, the test set's Pre, Rec and F1 drop to 0, indicating that all abnormal images are 
incorrectly detected as normal. This suggests that in the case of strong image noise, 
GasHis-Transformer tends to predict a more likely to normal category. In summary, for 
conventional noises, GasHis-Transformer has strong robustness to both Erlang noise and uniform 
noise. Similarly, the model also has strong robustness of epsilon between 0 and 0.1 for Gaussian, Rayleigh and Salt \& Pepper noise.

\vspace{-0.3cm}
\subsection{Extended Experiment}
Firstly, an extended experiment for gastrointestinal cancer detection is performed using additional 620 gastrointestinal images. Then, illustrative experiments are performed on a publicly available breast cancer dataset BreakHis and a lymphoma dataset immunohistochemical (IHC) stained lymphoma histopathological image dataset (IHC-LI-DS). The experimental setup of the extended experiments is generally based on the gastric cancer dataset and slightly adjusted with their respective characteristics. Finally, repeatability experiments are performed to demonstrate the stability of GasHis-Transformer.

\vspace{-0.2cm}
\subsubsection{Extended Experiment for Gastrointestinal Cancer Detection}
\begin{figure}[ht]
\centering
\includegraphics[trim={0cm 0cm 0cm 0cm},clip,width= 0.7\textwidth]{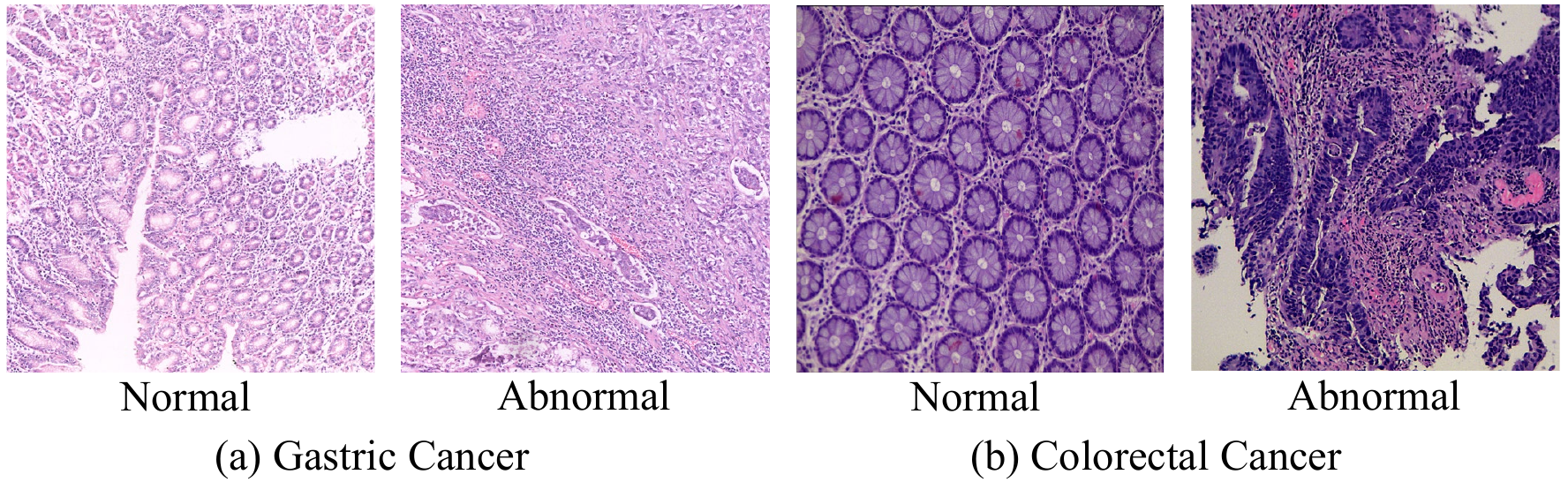}
\vspace{-0.5cm}
\caption{Some examples of gastrointestinal histopathological image.}
\label{fig:Contrast-GasIn}
\end{figure}

Gastrointestinal cancer includes gastric cancer and colorectal cancer. Due to gastric and colorectal organs have glands, their histopathological images have many similar features. Some examples of gastrointestinal cancer histopathological images is shown in Fig.~\ref{fig:Contrast-GasIn}. This extended experiment is used to evident that not only does the GasHis-Transformer model has outstanding performance in the GHID task, but it also has an excellent performance in the gastrointestinal cancer detection task. Based on the main experiment in Sub-section~\ref{subsubsection:er}, medical doctors often focus more on detecting abnormal categories. If images detect as abnormal by deep learning models, doctors need to conduct operations such as staging benign and malignant lesions, determining the area of the lesion, and determining whether it has spread extensively. Therefore,  doctors frequently prefer models with high detection rates in the abnormal category in clinical applications.

In the gastrointestinal cancer detection task, it includes gastric cancer detection and colorectal cancer detection. In the gastric cancer detection, training set, validation set and model parameters are followed the main experiment. The test data use the remaining 420 abnormal images in the dataset. In the colorectal cancer detection, the colorectal dataset contains 800 images of two different categories including abnormal category and normal category with image-level labels, which are provided by a medical doctor from China Medical University. 800 images are randomly assigned to training, validation and test sets with a ratio of 1: 1: 2 and the training set is expanded to six times. The model parameters are obtained by training GasHis-Transformer on the training set. Above all, 620 gastrointestinal cancer images including 420 gastric cancer images and 200 colorectal cancer images are used to test GasHis-Transformer. The gastrointestinal cancer detection results are shown in Table~\ref{table:GICI-Task}: For gastric cancer images, 409 images are correctly detected by the model with only 11 images are not detected, and Acc of the model for gastric cancer images reaches 97.97\%; for colorectal cancer images, 196 images are detected by the model with only 4 images not detected, and Acc of the model for colorectal images reaches 98.00\%. In summary, for 620 gastrointestinal images we have 600 detected images and 20 undetected images with a detection Acc of 97.58\%.
\begin{table}[h]
		\tiny
		\centering
		\caption{The result in the gastrointestinal cancer detection task. ([In \%].)}
	\label{table:GICI-Task}
	\vspace{-0.3cm}
	\begin{tabular}{|c|c|c|c|c|}
		\hline
		\multicolumn{2}{|c|}{Cancer Type}                            & Correct & Incorrect & Acc  \\ \hline
		\multirow{2}{*}{Gastrointestinal} & Gastric     & 409            & 11 & 97.97 \\ \cline{2-5} 	
	                                      & Colorectal  & 196            & 4                & 98.00 \\ \hline
		\multicolumn{2}{|c|}{Sum}                                    & 605 & 15               & 97.58 \\ \hline
	\end{tabular}
\end{table}

\vspace{-0.5cm}
\subsubsection{Extended Experiment for Breast Cancer Image Classification}
\begin{figure}[ht]
\centering
\includegraphics[trim={0cm 0cm 0cm 0cm},clip,width= 0.9\textwidth]{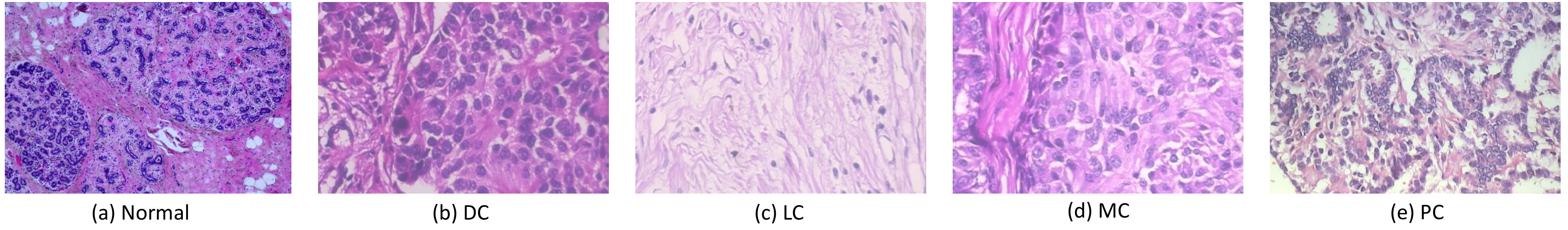}
\vspace{-0.5cm}
\caption{An example of $200 \times$ BreakHis Images.}
\label{fig:Example-Break}
\end{figure}

Breast cancer is associated with a high mortality rate in comparison with other cancers. We further demonstrate the well-performance of GasHis-Transformer in breast cancer image classification using BreakHis dataset~\cite{spanhol2015ADFBC}. In this paper, malignant tumors with a magnification of $200 \times$ are used for the four classifications including ductal carcinoma (DC), lobular carcinoma (LC), mucinous carcinoma (MC) and papillary carcinoma (PC) of the breast. An example of $200 \times$ BreakHis images is shown in Fig.~\ref{fig:Example-Break} and the data setting is shown in Table~\ref{table:9}. 

\begin{table}[h]
		\tiny
		\centering
		\caption{Data setting of BreakHis dataset for training, validation and test sets.}
	\label{table:9}
	\vspace{-0.3cm}
	\begin{tabular}{|c|c|c|c|c|c|}
	\hline
	\multicolumn{2}{|c|}{Image Type}      & Training & Validation & Test & Sum \\ \hline
	\multirow{2}{*}{DC}   & Oringin   & 538 & 179 & 179 & 896 \\ \cline{2-6} 
	                          & Augmented & 3228 & 1074 & 1074 & 5376 \\ \hline
	\multirow{2}{*}{LC} & Oringin   & 98 & 33 & 32 & 163 \\ \cline{2-6} 
	                          & Augmented & 588 & 198 & 192 & 978 \\ \hline
	\multirow{2}{*}{MC}   & Oringin   & 118 & 39 & 39 & 196 \\ \cline{2-6} 
	                          & Augmented &  708 & 234 & 234 & 1176 \\ \hline
	\multirow{2}{*}{PC} & Oringin   & 81 & 27 & 27 & 135 \\ \cline{2-6} 
	                          & Augmented & 486 & 162 & 162 & 810 \\ \hline
	\end{tabular}
\end{table}

The same experimental parameter setting is used for the classification of BreakHis data as that for HE-GHI-DS. The best classification results of the traditional CNN models on the BreakHis dataset is VGG-16, by the four criteria including Pre, Rec, F1 and Acc , which are 81.32\%, 79.20\%, 79.86\% and 85.74\%, respectively. Compared with VGG-16, Pre, Rec, F1 and Acc of GasHis-Transformer are increased by 2.60\%, 3.96\%, 3.62\% and 2.36\%, respectively and Pre, Rec, F1 and Acc of LW-GasHis-Transformer are increased by  3.22\%, 3.79\%, 3.83\% and 2.19\%, respectively. It shows that GasHis-Transformer and LW-GasHis-Transformer have better image classification performance on the BreakHis dataset. A comparison in extended experiments using different models on the BreakHis test set is shown in Table~\ref{table:11}. The experimental results further demonstrate that GasHis-Transformer and LW-GasHis-Transformer is outstanding in the GHID tasks as well as in other H\&E histopathological image classification tasks.

\begin{table}[htbp!]
		\tiny
		\centering
		\caption{A comparison of image classification results on the BreakHis test set. ([In \%].)}
		\label{table:11}
		\vspace{-0.3cm}
		{
		\begin{tabular}{|c|c|c|c|c|}
			\hline
			Models & Pre & Rec & F1 & Acc\\\hline
			GasHis-Transformer & $83.92\pm1.71$ & $\bm{83.16\pm1.74}$ & $83.48\pm1.38$ & $\bm{88.10\pm0.83}$ \\\hline
			LW-GasHis-Transformer & $\bm{84.54\pm3.00}$ & $82.99\pm3.19$ & $\bm{83.69\pm3.09}$ & $87.93\pm2.05$ \\\hline
			Xception~\cite{chollet2017xception} & $79.24\pm1.35$ & $78.62\pm1.23$ & $78.84\pm1.10$ & $85.33\pm0.75$\\\hline
			ResNet-50~\cite{he2016deep} & $74.60\pm2.48$ & $76.88\pm1.78$ & $75.54\pm2.15$ & $82.66\pm1.47$\\\hline
			Inception-V3~\cite{szegedy2016rethinking} & $79.18\pm4.76$ & $79.84\pm2.60$ & $79.02\pm2.98$ & $84.62\pm1.93$\\\hline
			VGG-16~\cite{simonyan2014very} & $81.32\pm1.89$ & $79.20\pm1.79$ & $79.86\pm1.39$ & $85.74\pm1.26$\\\hline
			VGG-19~\cite{simonyan2014very} & $78.42\pm1.70$ & $77.96\pm3.78$ & $77.14\pm2.72$ & $84.39\pm1.29$\\\hline
			ViT~\cite{dosovitskiy2020image} & $74.28\pm1.89$& $76.26\pm1.09$ & $75.04\pm1.06$ & $82.16\pm1.13$\\\hline
			BotNet-50~\cite{srinivas2021bottleneck} & $79.20\pm2.39$ & $80.72\pm3.69$ & $79.50\pm2.97$ & $85.32\pm1.65$\\\hline
		\end{tabular}
		}
\end{table}

\vspace{-0.5cm}
\subsubsection{Extended Experiment for Lymphoma Image Classification}
\begin{figure}[htbp!]
\centering
\includegraphics[trim={0cm 0cm 0cm 0cm},clip,width= 0.9\textwidth]{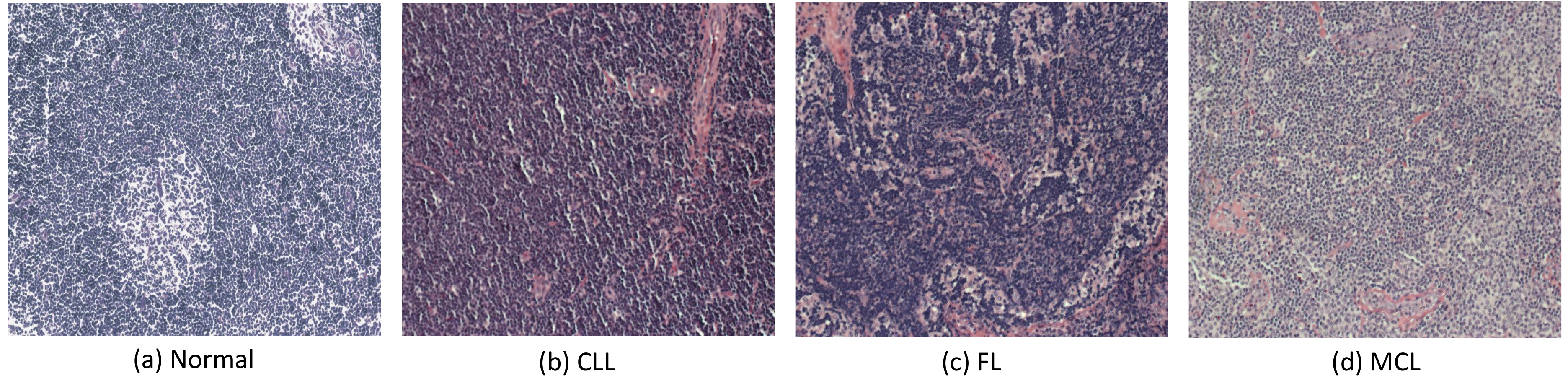}
\vspace{-0.5cm}
\caption{An example of IHC-LI-DS Images.}
\label{fig:Example-Lymphoma}
\end{figure}

Malignant lymphoma is one type of deadly cancer that affects the lymph nodes. Three types of malignant lymphoma are representative in the immunohistochemical (IHC) stained lymphoma histopathological image dataset (IHC-LI-DS)$\footnote{This dataset is open access on: Jaffe, E. and Orlov, N, NIA Intramural Research Program
Laboratory of Genetics, https://ome.grc.nia.nih.gov/iicbu2008/lymphoma/index.html}$: chronic lymphocytic leukemia (CLL), follicular lymphoma (FL) and mantle cell lymphoma (MCL). An example of IHC-LI-DS is shown in Fig.~\ref{fig:Example-Lymphoma}. A total of 374 images are available in IHC-LI-DS and the data setting is shown in Table~\ref{table:6}. Since three different types of lymphoma are classified according to the shape of lymphocytes, directly resizing the image of lymphoma makes it challenging to distinguish the small and dense lymphocytes. Therefore, we crop the whole lymphoma image into patches of $224 \times 224$ pixels as the standard input to GasHis-Transformer and LW-GasHis-Transformer, which augment the datasets by 24 times.

\begin{table}[htbp!]
		\tiny
		\centering
		\caption{Data setting of IHC-LI-DS for training, validation and test sets.}
	\label{table:6}
	\begin{tabular}{|c|c|c|c|c|c|}
	\hline
	\multicolumn{2}{|c|}{Image Type}      & Training & Validation & Test & Sum \\ \hline
	\multirow{2}{*}{CLL}   & Oringin   & 40 & 30 & 43 & 113 \\ \cline{2-6} 
	                          & Augmented & 960 & 720 & 1032 & 2712 \\ \hline
	\multirow{2}{*}{FL} & Oringin   & 40 & 30 & 69 & 139 \\ \cline{2-6} 
	                          & Augmented & 960 & 720 & 1656 & 3336 \\ \hline
	\multirow{2}{*}{MCL}   & Oringin   & 40 & 30 & 52 & 122 \\ \cline{2-6} 
	                          & Augmented &  960 & 720 & 1248 & 2928 \\ \hline
	\end{tabular}
\end{table}

Table~\ref{table:8} summarizes the experimental results in the same experimental parameter setting of HE-GHI-DS to classify the three-class IHC-LI-DS. GasHis-Transformer has the best performance in Rec, F1 and Acc while LW-GasHis-Transformer has the best performance in Pre compared with the other models. In the IHC-LI-DS, the best performance of the traditional models is Xception, which has the highest Pre, Rec, F1 and Acc among the traditional CNN models, reaching 80.72\%, 80.58\%, 80.00\% and 81.48\%, respectively. However, the performance of GasHis-Transformer and LW-GasHis-Transformer is even better than that of Xception. Pre, Rec, F1 and Acc of GasHis-Transformer reach an outstanding 82.42\%, 83.30\%, 83.16\% and 84.34\% respectively while LW-GasHis-Transformer reach 82.66\%, 82.70\%, 82.38\% and 83.64\%, respectively. Therefore, GasHis-Transformer and LW-GasHis-Transformer not only have an excellent classification performance on H\&E stained datasets, but also do well in IHC stained datasets.
\begin{table}[!t]
		\tiny
		\centering
		\caption{A comparison of image classification results on the IHC-LI-DS test set. ([In \%].)}
		\label{table:8}
		\setlength{\tabcolsep}{2mm}
		\vspace{-0.3cm}
		{
		\begin{tabular}{|c|c|c|c|c|}
			\hline
			Models & Pre & Rec & F1 & Acc\\\hline
			GasHis-Transformer & $82.42\pm1.97$ & $\bm{83.30\pm1.37}$ & $\bm{83.16\pm1.02}$ & $\bm{84.34\pm0.72}$ \\\hline
			LW-GasHis-Transformer & $\bm{82.66\pm0.90}$ & $82.70\pm0.82$ & $82.38\pm0.63$ & $83.64\pm0.78$ \\\hline
			Xception~\cite{chollet2017xception} & $80.72\pm0.92$	& $80.58\pm0.74$ & $80.00\pm0.98$ & $81.48\pm1.12$ \\\hline
			ResNet-50~\cite{he2016deep} & $77.66\pm0.81$	& $78.06\pm0.87$ & $77.36\pm0.81$ & $78.58\pm0.77$\\\hline
			Inception-V3~\cite{szegedy2016rethinking} & $78.26
\pm1.35$	& $78.78\pm1.28$ & $78.22\pm1.49$ & $79.17\pm1.46$\\\hline
			VGG-16~\cite{simonyan2014very} & $76.48\pm0.71$	& $77.00\pm0.76$ & $76.58\pm0.73$ & $77.81\pm0.68$\\\hline
			VGG-19~\cite{simonyan2014very} & $77.04\pm1.70$	& $77.34\pm1.56$ & $76.78\pm1.66$ & $77.91\pm2.18$\\\hline
			ViT~\cite{dosovitskiy2020image} &$73.24\pm1.90$	& $74.10\pm1.76$ & $73.12\pm1.98$ & $74.33\pm2.03$\\\hline
			BotNet-50~\cite{srinivas2021bottleneck} & $77.54\pm2.39$	& $76.86\pm2.12$ & $76.78\pm3.00$ & $77.20\pm2.40$\\\hline
		\end{tabular}
		}
\end{table}

%

\vspace{-0.5cm}
\subsection{Experimental Environment and Computational Time}
A workstation with Windows 10, AMD Ryzen 7 4800HS 2.90GHz, GeForce RTX 2060 6GB and 16 GB RAM is utilized in the experiment. Matlab R2020b is used to do image pre-processing. Python 3.6, Pytorch 1.7.0 and torchvision 0.8.0 are used for deep learning. Table~\ref{table:time} shows the parameter size and training time on three datasets of eight deep learning models. It takes 0.86 and 0.67 hours to train GasHis-Transformer and LW-GasHis-Transformer with 840 training and validation images in 75 epochs, and only 30 Sec to test GasHis-Transformer and LW-GasHis-Transformer on 840 images (0.036 Sec per image).
\begin{table}[htbp!]
		\tiny
		\centering
		\caption{The parameter size (MB) and training time (hour) in all experiments.}
		\label{table:time}
		\vspace{-0.3cm}
		{
		\begin{tabular}{|c|c|c|c|c|c|}
			\hline
\multirow{2}{*}{Models} & \multirow{2}{*}{Parameter Size} & \multicolumn{3}{l|}{Training Time} \\ \cline{3-5} 
                        &                                  & Gastry  & Breast  & Lymphoma \\ \hline
			GasHis-Transformer & 155 & 0.86 & 2.78 & 2.39 \\\hline
			LW-GasHis-Transformer & 77 & 0.67 & 2.14 & 1.60 \\\hline            
			Xception~\cite{chollet2017xception} & 79 & 0.75 & 2.36  & 1.75  \\\hline
			ResNet-50~\cite{he2016deep} & 90 & 0.69 & 1.73 & 1.46 \\\hline
			Inception-V3~\cite{szegedy2016rethinking} & 83 & 0.73 & 1.58 & 1.58 \\\hline
			VGG-16~\cite{simonyan2014very} & 268 & 0.78 & 2.39 &  1.78 \\\hline
			VGG-19~\cite{simonyan2014very} & 298 & 0.81 & 2.76 & 2.01 \\\hline
			BotNet-50~\cite{srinivas2021bottleneck} & 72  & 0.69 & 1.72 & 1.46 \\\hline
			ViT~\cite{dosovitskiy2020image} & 48 & 0.69 & 1.58 &  1.29 \\\hline
		\end{tabular}
		}
\end{table}

\vspace{-0.2cm}
\section{Conclusion and Future Work}
\label{section:c}
\vspace{-0.3cm}
We have proposed a GasHis-Transformer model and its lightweight version called LW-GasHis-Transformer to detect gastric cancer in histopathological images. This approach combines the advantages of the classical CNN model in extracting local information and uses the most recent Transformer model in capturing long-range correlation considering the global and local associations of images in a unified context. In the experiments, GasHis-Transformer and LW-GasHis-Transformer are tested on a gastric cancer histopathological dataset with accuracies of 97.97\% and 96.43\%, respectively, showing their potential in the GHID tasks. Also, we have conducted a gastrointestinal cancer detection task to demonstrate that GasHis-Transformer and LW-GasHis-Transformer have excellent cancer detection abilities. Finally, we have extended tests for classification tasks on a breast cancer dataset and a lymphoma dataset with excellent accuracies 88.10\% and 84.34\% for GasHis-Transformer and 87.93\% and 83.64\% for LW-GasHis-Transformer, respectively, demonstrating an encouraging generalizability both in H\&E stained and IHC stained histopathological images. Since a small dataset in this experiment may lead to a degradation of the classification performance of GasHis-Transformer model, we consider using methods such as domain adaptation and few-shot learning to solve this problem in future work.

\vspace{-0.5cm}
\section*{Acknowledgements and Conflict of Interest}
\vspace{-0.3cm}
This work is supported by National Natural Science Foundation of China (No. 61806047). We thank Miss Zixian Li and Mr. Guoxian Li for their important discussion. We also thank Mr. Jinghua Zhang for his contribution to the revision of this paper. There is no conflict of interest in this paper.

\vspace{-0.5cm}
\bibliographystyle{elsarticle-num} 
\footnotesize
\bibliography{Haoyuan}

\end{document}